\documentclass[conference]{IEEEtran}
\IEEEoverridecommandlockouts
\usepackage{cite}
\usepackage{amsmath,amssymb,amsfonts}
\usepackage{bbm}
\usepackage{algorithm}
\usepackage{algorithmic}
\usepackage{graphicx}
\usepackage{subfigure}
\usepackage{textcomp}
\usepackage{xcolor}
\usepackage{pifont}
\usepackage{multirow}
\def\BibTeX{{\rm B\kern-.05em{\sc i\kern-.025em b}\kern-.08em
    T\kern-.1667em\lower.7ex\hbox{E}\kern-.125emX}}
\begin{document}

\title{FedPaI: Achieving Extreme Sparsity in Federated Learning via Pruning at Initialization
}


 \author{\IEEEauthorblockN{Haonan Wang\textsuperscript{*}}
 \IEEEauthorblockA{\textit{USC Information Sciences Institute} \\
 \textit{University of Southern California}\\
 Marina Del Rey, CA, USA \\
 \texttt{haonanwa@usc.edu}}
 \and
 \IEEEauthorblockN{Zeli Liu\textsuperscript{*}}
 \IEEEauthorblockA{\textit{Viterbi School of Engineering} \\
 \textit{University of Southern California}\\
 Los Angeles, CA, USA \\
 \texttt{zeliliu@usc.edu}}
 \and
 \IEEEauthorblockN{Kajimusugura Hoshino}
 \IEEEauthorblockA{\textit{Viterbi School of Engineering} \\
 \textit{University of Southern California}\\
 Los Angeles, CA, USA \\
 khoshino@usc.edu}
 \and
 \IEEEauthorblockN{Tuo Zhang}
 \IEEEauthorblockA{\textit{Viterbi School of Engineering} \\
 \textit{University of Southern California}\\
 Los Angeles, CA, USA \\
 tuozhang@usc.edu}
 \and
 \IEEEauthorblockN{John Paul Walters}
 \IEEEauthorblockA{\textit{USC Information Sciences Institute} \\
 \textit{University of Southern California}\\
 Arlington, VA, USA \\
 \texttt{jwalters@isi.edu}}
 \and
 \IEEEauthorblockN{Stephen Crago}
 \IEEEauthorblockA{\textit{USC Information Sciences Institute} \\
 \textit{University of Southern California}\\
 Arlington, VA, USA \\
 \texttt{crago@isi.edu}}
 }

\maketitle

\begin{abstract}
Federated Learning (FL) enables distributed training on edge devices but faces significant challenges due to resource constraints in edge environments, impacting both communication and computational efficiency. Existing iterative pruning techniques improve communication efficiency but are limited by their centralized design, which struggles with FL's decentralized and data-imbalanced nature, resulting in suboptimal sparsity levels. To address these issues, we propose FedPaI, a novel efficient FL framework that leverages Pruning at Initialization (PaI) to achieve extreme sparsity. FedPaI identifies optimal sparse connections at an early stage, maximizing model capacity and significantly reducing communication and computation overhead by fixing sparsity patterns at the start of training.
To adapt to diverse hardware and software environments, FedPaI supports both structured and unstructured pruning. Additionally, we introduce personalized client-side pruning mechanisms for improved learning capacity and sparsity-aware server-side aggregation for enhanced efficiency. Experimental results demonstrate that FedPaI consistently outperforms existing efficient FL that applies conventional iterative pruning with significant leading in efficiency and model accuracy. For the first time, our proposed FedPaI achieves an extreme sparsity level of up to 98\% without compromising the model accuracy compared to unpruned baselines, even under challenging non-IID settings. By employing our FedPaI with joint optimization of model learning capacity and sparsity, FL applications can benefit from faster convergence and accelerate the training by 6.4 to 7.9$\times$.
\end{abstract}

\begin{IEEEkeywords}
Federated Learning, efficient, pruning, sparsity
\end{IEEEkeywords}

\section{Introduction}
Federated Learning (FL)~\cite{mcmahan2017communication, mcmahan2017fedavg} has emerged as a promising approach for decentralized machine learning on edge devices, which are rapidly growing in number and capability. As data generated by these devices increases, traditional centralized training methods face significant limitations, especially in applications where data privacy is critical. FL allows multiple devices to collaboratively train a shared model without needing to transfer sensitive data to a central server, thus preserving user privacy while utilizing the diverse data spread across these devices.
However, FL faces significant challenges, particularly in managing the escalating communication and computation costs associated with frequent model updates. As machine learning models evolve from CNNs~\cite{simonyan2014very, he2016deep, 9195240, 9401765} to more complex architectures like Transformers~\cite{vaswani2017attention, dosovitskiy2020image, 10096223}, their size has grown substantially, demanding increasingly massive resources for training, even in centralized data centers. Consequently, these challenges are especially acute for FL in edge environments, since most existing commercial edge devices only possess limited computing and bandwidth resources. 

To enhance machine learning efficiency, researchers have extensively investigated model compression techniques~\cite{han2016deep, dong2019hawq, 10096632, wang2024moq}, with pruning~\cite{frankle2018the, 8683512} emerging as a particularly effective approach for reducing model size and computational requirements. A landmark discovery in this field, the Lottery Ticket Hypothesis (LTH)~\cite{frankle2018the}, demonstrated that dense neural networks inherently contain sparse subnetworks that can achieve comparable test accuracy to their dense counterparts when trained from the same initialization. Building on these insights, recent FL works~\cite{9708944, wu2023efficient, shi2024efficient} have extended pruning techniques to federated learning environments to address communication bottlenecks.

However, the direct application of centralized pruning methods to FL reveals significant limitations in achieving extreme sparsity. While iterative pruning~\cite{han2016deep} has shown remarkable success in centralized scenarios, achieving sparsity levels exceeding 90\% without compromising model performance, state-of-the-art (SOTA) communication-efficient FL approaches~\cite{9708944, wu2023efficient, shi2024efficient, Feng2024ModalityMirrorIA} have yet to match these sparsity levels. This performance gap stems from the fundamental mismatch between conventional pruning methods-designed for centralized infrastructure with independently and identically distributed (IID) data—and the unique characteristics of FL environments. The decentralized nature of FL introduces two critical challenges: First, pruning strategies must balance efficient distributed communication and computation while maintaining model accuracy. Second, the presence of non-identically independently distributed (non-IID) data across clients necessitates personalized pruning strategies, making it challenging for traditional centralized pruning methods to effectively and efficiently aggregate these diverse sparse structures during model fusion.

To address these issues of existing pruning and further improve the efficiency of the FL system, in this paper, we introduce FedPaI, an efficient FL framework designed to enhance both system efficiency through extreme sparsity and model performance via pruning at initialization (PaI) scheme. Unlike traditional pruning methods that gradually increase model sparsity during training, FedPaI uses PaI to identify optimal sparse connections at the very beginning of the training process. This approach leverages gradient information to retain the model’s capacity and fixes the sparsity at an early stage, reducing the need for repeated pruning and minimizing communication overhead. The optimal sparse connection pattern found by PaI methods is determined by maximizing the gradient flow, which preserves the learning ability of the pruned network to the greatest extent, while magnitude-based weight pruning methods, e.g., LTH~\cite{9708944}, are not able to retain such learning capability and usually result in failure of training convergency, especially under a high sparsity ratio.
Our main contributions are summarized as follows:
\begin{itemize}
    \item To the best of our knowledge, we are the first to introduce PaI to FL and design a communication- and computation-efficient FL framework (FedPaI). 
    \item For the first time, FedPaI achieves over 98\% pruning rate on popular CNN models without harming the model performance, even under an extreme non-IID setting.
    \item FedPaI is designed to be a flexible FL framework, providing both structured and unstructured schemes. This ensures that our framework can adapt to the diverse platforms across different hardware, achieving significant performance and efficiency gains no matter whether dedicated hardware support is available. It also offers users flexibility with choices of server- and client-side pruning to accommodate various applications with different upload/download bandwidth quotas.
    \item Experimental results show that the proposed FedPaI significantly outperforms the existing efficient FL frameworks in terms of both efficiency and model accuracy, enabling acceleration for FL applications by 6.4$\times$ to 7.9$\times$.
\end{itemize}

\section{Background and Motivation}
\subsection{Model pruning}\label{sec:intro_pruning}
Neural network pruning has emerged as a crucial technique for reducing model complexity and improving deployment efficiency in deep learning systems. Traditional iterative magnitude-based pruning, pioneered by Han et al.~\cite{han2016deep}, progressively removes weights with small magnitudes during training, requiring multiple iterations of pruning and fine-tuning to maintain accuracy. The seminal Lottery Ticket Hypothesis (LTH) revealed that dense neural networks contain sparse subnetworks (winning tickets) that can be trained in isolation to achieve comparable or even superior performance to the original network. However, these iterative approaches are computationally intensive, often requiring multiple training cycles. To address this limitation, Early Bird Ticket~\cite{you2019drawing} was proposed, demonstrating that winning tickets can be identified in the early stages of training, significantly reducing the computational overhead of traditional iterative pruning methods. More recently, pruning at initialization (PaI) methods have gained attention for their ability to identify optimal sparse architectures before training begins. Methods like SNIP (Single-shot Network Pruning)~\cite{lee2019snip} evaluate connection sensitivity using gradients from a single backward pass, while GraSP (Gradient Signal Preservation)~\cite{tanaka2020pruning} focuses on preserving gradient flow by analyzing the interaction between weights and gradients at initialization. These PaI approaches offer significant advantages in computational efficiency by eliminating the need for iterative pruning cycles while maintaining competitive performance.

\subsection{Efficient Federated Learning}
Federated Learning (FL) enables collaborative model training across distributed edge devices while preserving data privacy. To address the resource constraints in FL deployment, recent works~\cite{9708944, wu2023efficient, shi2024efficient} have incorporated model pruning techniques to reduce communication and computation overhead. These approaches typically adopt iterative magnitude-based pruning, where weights with small magnitudes are progressively removed during training, following the methodology of LTH~\cite{frankle2018the}. However, such centralized pruning strategies face fundamental limitations in federated settings. First, generating identical sparsity patterns across clients fails to accommodate the personalized features inherent in non-IID data distributions. Second, magnitude-based criteria provide limited insight into gradient flow, which is crucial for model optimization—--a limitation that becomes particularly severe in FL where non-IID data already challenges model convergence. To address these limitations, we propose that PaI methods offer a promising alternative to identify optimal sparse architectures before training begins, potentially enabling more efficient and effective model sparsification in federated settings.

\begin{algorithm}[t]
\caption{Unstructured Pruning at Initialization}
\begin{algorithmic}[1]\label{alg:PaI}
\REQUIRE Loss function $L$, training dataset $\mathcal{D}$, sparsity $\kappa$;
\STATE $\mathbf{W} \leftarrow$ WeightInit$(\mathbf{W})$ 
\STATE $\mathcal{D}^b = \{(x_i, y_i)\}_{i=1}^b \sim \mathcal{D}$ 
\STATE $\mathbf{s} \leftarrow$ ImportanceScore$(grad(L(\mathbf{W}); \mathcal{D}^b))$ 
\STATE $\tilde{s}_\kappa \leftarrow$ DescendingSort$(\mathbf{s};\kappa)$ 
\STATE $\mathbf{m} \leftarrow \mathbbm{1} [\mathbf{s} - \tilde{s}_\kappa \geq 0]$ 
\STATE $\mathbf{W}^* \leftarrow \mathbf{m} \odot \mathbf{W}$ 
\STATE $\mathbf{W}^* \leftarrow \arg \min_{\mathbf{W}^* \in \mathbb{R}^m} L(\mathbf{W}^*; \mathcal{D})$ 
\vspace{1em}
\STATE \textbf{Function} GraSPImportanceScore$(L(\mathbf{W});\mathcal{D}^b)$:
\STATE \hspace{1em} $\mathbf{g} = grad(L(\mathbf{W});\mathcal{D}^b)$
\STATE \hspace{1em} $\mathbf{Hg} = grad(\mathbf{g}^\top stop\_grad(\mathbf{g}); \mathcal{D}^b)$
\STATE \hspace{1em} $\mathbf{s} \leftarrow -\mathbf{W} \odot \mathbf{H}\mathbf{g}$
\STATE \hspace{1em} \textbf{return} $\mathbf{s}$
\end{algorithmic}
\end{algorithm}

\section{System design of FedPaI}
In this work, we mainly focus on exploring both unstructured and structured PaI pruning techniques to improve the performance and efficiency of FL. We consider the federated learning setting that is similar to vanilla FedAvg~\cite{mcmahan2017fedavg}, in which the FL system is composed of a server with N clients, whose data is only locally kept without sharing. In the following, we will depict how to accommodate PaI methods to achieve high sparsity and design an efficient FL system that can fully harness sparsity for training acceleration.


\subsection{Efficient Federated Learning Paradigm via Pruning}
In general, the key idea is to leverage pruning methods to sparsify either the local or global model, so that only sparse parameters are transmitted between clients and the server, improving the communication efficiency of the FL system. 

Specifically, there is a global model $W_g$ maintained by the server and a local model $W_{c, i}$ kept by client $C_i$ from the available client set $\mathcal{C}= \{ C_1, \ldots, C_N \}$. Each client possesses its own part of the local dataset $D_i \subset \mathcal{D}$, where $\mathcal{D}$ is the collection of all training data. In the efficient FL paradigm, each client will maintain a local mask $m_i \in \{0,1\}^{|W_i|}$ to prune less important connections. This mask can be learned by the client or the server. During the $t$-th round of the training iteration, a randomly selected set of clients $\mathcal{S}_t \subset \mathcal{C}$ will participate in the learning, and each client learns from its local set of data $D_i$ and updates the local model $W_{c,i}$:
\begin{equation}\label{eq:cli_mask}
    W_{c,i}(t) = W_{c,i}(t-1) - \eta \nabla \mathcal{L}(W_{c,i}(t-1) \odot m_i)
\end{equation}
where $\mathcal{L}(\cdot)$ represents the loss of the network, and $\eta$ denotes the learning rate. After all clients finish local updates, the server will perform the averaging procedure of FedAvg over all pruned local models $W_{c,i}$ of clients $C_i \in \mathcal{S}_t$ :
\begin{equation}\label{eq:ori_agg}
    W_g(t) = \sum_{c_i \in \mathcal{S}}W_{c,i}(t) \odot m_i / |\mathcal{S}_t|
\end{equation}
Masks $m_i$ are applied to the local models to obtain sparse weights, so that the upload bandwidth can be saved. The server also maintains a mask version $m_g$. After the global model $W_g$ gets updated, the server will prune it with the global mask $m_g$ and distribute the pruned model to all clients:
\begin{equation}\label{eq:glo_mask}
    W_{c,i}(t+1) = W_g(t) \odot m_g \text{, for } C_i \in \mathcal{C}
\end{equation}
Therefore, the download link can also be saved with a sparse representation of the global weight.
From Eqs.\ref{eq:cli_mask} and~\ref{eq:glo_mask}, the key step for designing an efficient FL system is determining the pruning method for generating the mask $m_i$ and $m_g$. Hence, we will analyze the existing pruning methods, and answer the question: \textit{what pruning method can best accommodate the FL paradigm?}

\subsection{Unstructured PaI for personalized learning}
First, we particularly investigate the unstructured PaI method, due to its powerful capability to maintain learning capacity under high sparsity. We specifically refer to the state-of-the-art GraSP pruning as \textit{\textbf{the unstructured PaI}} in the following sections. The general procedure of the unstructured PaI for centralized training is illustrated in Algorithm~\ref{alg:PaI}. It will first sample a batch of data (usually the batched samples of the first training epoch) and run one forward and backward propagation to collect necessary gradient information. Then, it will evaluate the importance of the weights based on the gradients. An advanced \textbf{\textit{$ImportanceScore(\cdot)$}} is defined in GraSP by approximately measuring the impact on the gradient flow via the Hessian-gradient product (line 8-12 in Algorithm~\ref{alg:PaI}).

\begin{figure}
    \centering
    \includegraphics[width=1\linewidth]{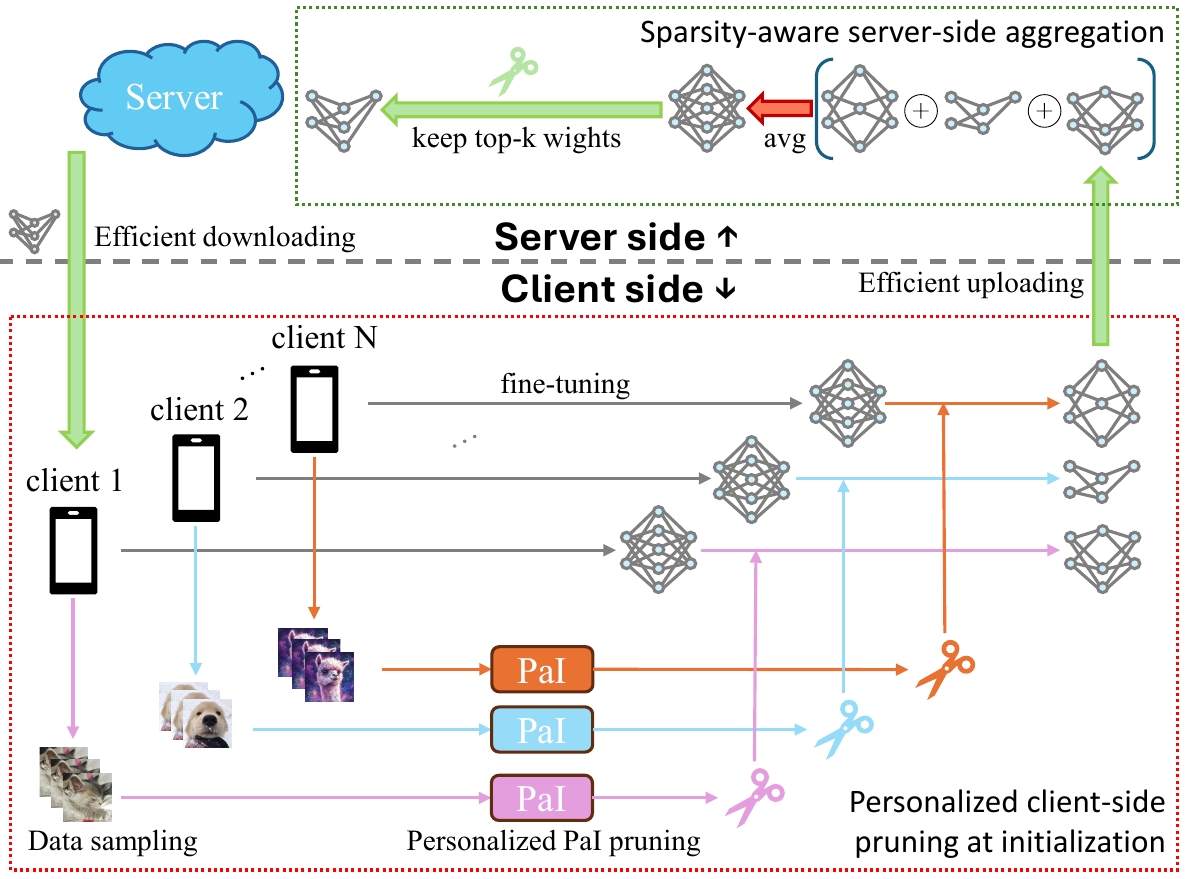}
    \caption{FedPaI system with personalized client-side unstructured PaI (FedPaI-U).}
    \label{fig:fedpai-system-unstruct}
\end{figure}

However, it is non-trivial to fully exploit the potential of PaI for efficiency improvement, we explore a comprehensive design space and propose the following approaches to accommodate PaI to FL. An overview of the proposed unstructured-PaI-based FadPaI system is shown in Figure~\ref{fig:fedpai-system-unstruct}.

\begin{figure}
        \centering
        \includegraphics[width=1\linewidth]{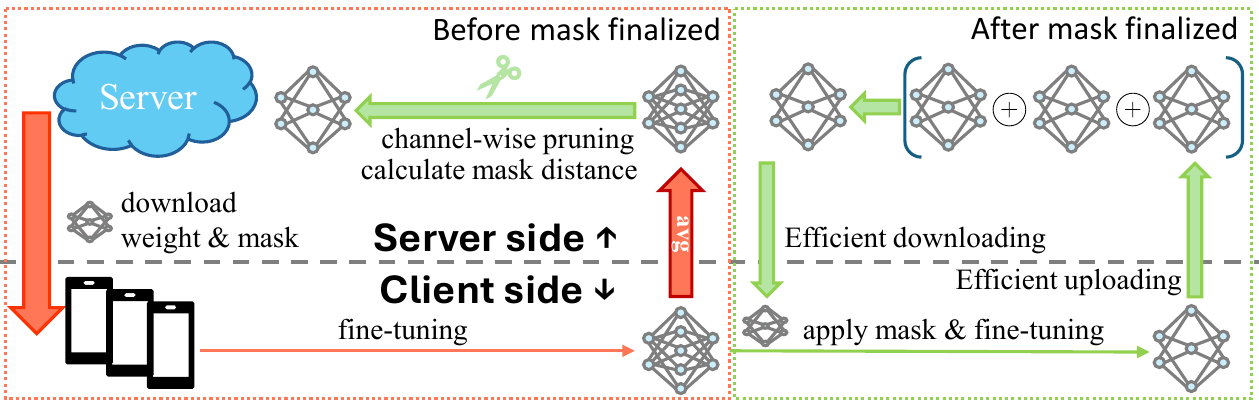}
        \caption{{FedPaI system with server-side structured PaI (FedPaI-S).}}
        \label{fig:fedpai-system-struct}
    \end{figure}

\textbf{\textit{Personalized client-side pruning at initialization.}} In FL, there is a global model lying on the server side and multiple local models located on the client side, and PaI can be applied to generate either mask $m_g$ (server-side) or $m_{i,i\in\mathcal{C}}$ (client-side). Most existing efficient FL works only consider server-side pruning. For example, Fed-LTP~\cite{shi2024efficient} only applies pruning on the server side and then distributes the global mask to all clients. Wu et al.~\cite{wu2023efficient} argue that the aggregation operation on the server will cancel out the sparsity brought by client-side pruning, making the global model dense again and wasting the download link. Thus, they also choose server-side pruning. However, considering that different clients possess different data distributions, server-side pruning neglects the possibility of learning personalized sparse connections for each client. In contrast, operating data sampling of PaI on the client side (line 2 in Algorithm~\ref{alg:PaI}) could better mitigate the model drift caused by the non-IID client data distribution nature of FL. Specifically, it generates more personalized masks for each client compared to previous approaches, which improves the local model learning capability, especially under a high sparsity ratio.
\textit{Thus, we propose applying personalized client-side PaI in our proposed FedPaI system to improve model learning performance}. We implement both server- and client-side PaI in the FedPaI system and make a more detailed comparison and analysis in Section.~\ref{sec:experiments}.

\textbf{\textit{Sparsity-aware server-side aggregation.}} As shown in Figure~\ref{fig:fedpai-system-unstruct}, the pruned weight will be non-zero again after the average aggregation as long as one client is viewing it as non-zero, so the naive aggregation mechanism of FedAvg will cancel out the sparsity obtained by the client-side PaI. To resolve this client-side sparsity cancellation issue, while still leveraging the capability of personalized learning of client-side PaI, \textit{we propose a sparsity-aware server-side aggregation mechanism in our FedPaI system.} We hypothesize that the weight magnitude of a specific client can be interpreted as the importance of the connection from the perspective of that client, and thus, the global model obtained from the average aggregation represents a weighted average of weight importance. Then, to maintain the efficiency of the download link when distributing the global model to all clients, we only keep the top-$\kappa$ important global weight based on its magnitude, where $1-\kappa$ is the target pruning ratio. Thus, in our FadPaI, the aggregation in Eq.~\ref{eq:ori_agg} will be modified as:
\begin{equation}\label{eq:spr_agg}
    W_g(t) = \text{Top-}\kappa(\sum_{c_i \in \mathcal{S}}W_{c,i}(t) \odot m_i / |\mathcal{S}_t|)
\end{equation}
This sparsity-aware server-side aggregation mechanism can preserve the most important connections learned by clients, while still maintaining an efficient downloading bandwidth with client-side pruning.

\subsection{Structured PaI for computation acceleration}
In the pursuit of an efficient FL system, reducing resource requirements and accelerating training are paramount goals. However, real-world FL deployments often involve client devices that are standard commercial-grade hardware, lacking high-performance chips capable of handling intensive computations. Although pruning techniques can effectively reduce communication and memory overhead, their impact on computational acceleration remains limited unless specialized hardware supporting sparse matrix operations is available. Most exciting works~\cite{9708944, wu2023efficient, shi2024efficient} only consider unstructured pruning and overlook the potential of harnessing sparsity for computation acceleration. To address this challenge, we extend our proposed FedPaI framework to incorporate both structured and unstructured pruning methods. This flexibility allows FedPaI to seamlessly adapt to varying infrastructures with or without sparsity-compatible hardware support, empowering users to choose the pruning approach that best aligns with their efficiency and accuracy constraints.

Specifically, we leverage a state-of-the-art channel-wise structured pruning method, \textit{\textbf{Early Bird Ticket (EBT)}}~\cite{you2019drawing}, as part of the FedPaI framework. In the subsequent sections, we refer to EBT as \textit{\textbf{the structured PaI}}. By removing unimportant channels in a structured manner and regrouping the remaining channels into a smaller model, structured PaI enables even standard devices without sparsity-aware hardware to benefit from accelerated training. This is achieved by significantly reducing the computational demand of the smaller model and minimizing communication overhead. Similar to unstructured PaI, structured PaI identifies and fixes the sparsity pattern at an early stage of training. Notably, instead of adopting a progressive pruning and training approach, EBT finalizes the mask pattern based on an early-stopping mechanism:
\begin{equation}
    m^* = m(t) \text{ if } HammingDistance(m(t), m(t-1)) < \epsilon
\end{equation}
where $m^*$ denotes the fixed sparse connection, and it is fixed when the distance of masks of two consecutive iterations is smaller than a threshold $\epsilon$. Despite being different from the gradient flow criterion used in unstructured PaI, structured PaI also effectively preserves the learning capacity of the pruned model. This is because it aggregates connection-relevant information over multiple epochs, implicitly capturing gradient flow when stabilizing the sparsity pattern. Furthermore, as channel-wise pruning operates on a coarse-grained level, it induces minimal variations in sparsity patterns across clients' personalized data. Consequently, structured PaI is applied directly on the server side, where only one global mask is generated and distributed to all clients, ensuring a seamless and efficient training process.

\subsection{Analysis of FedPaI}
By jointly optimizing the pruning and the FL paradigm from both the client and server sides, our proposed FedPaI shows several advantages.

\textit{\textbf{Better aggressive-pruning learning capacity}}. We noticed that conventional pruning methods that are directly applied in existing efficient FL works fail to achieve similar sparsity as their counterparts in a centralized training scenario~\cite{9708944,frankle2018the}. We assume it is because the magnitude-based pruning fails to preserve the learning capability of the pruned model, especially under high sparsity. In contrast, the gradient-based pruning criterion employed in unstructured PaI (line 8 in Alg.~\ref{alg:PaI}) is designed to maximize the gradient flow for better learning capacity, and structured PaI can also effectively preserve learning capacity via stabilizing sparse pattern over multiple epochs. We will show in Sec.~\ref{sec:experiments} that both structured and unstructured FedPaI can illustrate great model performance over conventional efficient FL frameworks, especially under an extremely high pruning rate.

\textit{\textbf{Better personalization adaptiveness}}. Given that PaI can better capture each client's feature via data sampling (line 2 of Algo.~\ref{alg:PaI}) and generate a personalized sparsity pattern that specifically accommodates the distribution of the specific client; thus, we speculate that our client-side PaI mechanism can provide a better model performance, especially under an extreme non-IID setting.
    
\textit{\textbf{Better communication efficiency}}. 
Compared to conventional iterative pruning methods~\cite{han2016deep,9708944}, which progressively approaches the target pruning rate for many training iterations, PaI achieves the target sparsity ratio at initialization, so that it can save more communication resources. Moreover, we resolve the client-side sparsity cancellation issue via sparsity-aware server-side aggregation for client-side unstructured PaI so that the FedPaI system will not suffer from an inefficient download link. To sum up, our proposed FedPaI shows superior efficiency over existing efficient FL frameworks.

\section{Experimental Evaluation}\label{sec:experiments}

\subsection{Experimental Settings}
\textbf{\textit{Implementation.}} We implement the proposed FedPaI system using CPU/GPU-based distributed training on Nvidia A100 GPUs. We build our system code based on Pytorch version 2.1.2.


\textbf{\textit{Model and Data Heterogeneity.}} Following prior work~\cite{Zhang2023GPTFLGP}, we evaluate FedPaI on the CIFAR-10 dataset~\cite{cifar10} using VGG19~\cite{simonyan2014very} and ResNet18~\cite{he2016deep}. Besides the default IID data partitioning, we perform the Dirichlet non-IID sampling to simulate the real-world challenging as suggested in previous work~\cite{Alam2023FedAIoTAF}. We partition the dataset among $J$ clients by sampling $\mathbf{p}_k \sim \text{Dir}_J(\alpha)$ and allocating a proportion $\mathbf{p}_{k,j}$ of the training samples from class $k$ to client $j$, where $\text{Dir}_J(\alpha)$ denotes the Dirichlet distribution with concentration parameter $\alpha$. We fully evaluate the FedPaI covering a wide range of $\alpha$, from 0.1 to 1.0, to simulate both extreme non-IID and IID cases.

\textbf{\textit{FL training hyperparameters.}} Since we mainly focus on investigating the pruning in FL in this work, but not the FL paradigm itself, we employ the same hyperparameters for all experiments FL unless explicitly stated. In our experiments, 10\% of a total of 100 clients will be randomly activated each communication round to participate in the training process. Each selected client performs local training on its private data for 10 epochs. The initiated learning rate is 0.1, and we schedule it to decrease by $10\times$ at epoch 400.

\textbf{\textit{Baselines.}} For accuracy evaluation, we train baseline models from scratch with the native FedAvg scheme and report their accuracy as baselines. Besides, we select LotteryFL~\cite{shi2024efficient} as our baseline efficient FL framework, as it explores pruning methods within the FL setting. Specifically, LotteryFL employs the LTH~\cite{frankle2018the} strategy, which follows the conventional iterative pruning approach. For a fair comparison, we independently implement LotteryFL using the same training hyperparameter settings, but with a fixed learning rate to 0.1 which aligns to its best practice. In particular, we adopt the same Dirichlet non-IID sampling scheme as in our experiments. This is because LotteryFL employs a customized 2-class non-IID sampling strategy, where each client is assigned data from only two specific classes. Although this approach aligns with their experimental setup, it is highly specific and does not generalize well to broader federated learning scenarios. Additionally, it is important to note that the results reported in LotteryFL are based on training accuracy evaluated on these 2-class private datasets, which naturally leads to an inflated accuracy compared to a more general non-IID setting.

\textbf{\textit{Experiment annotation.}} In our experimental design, we denote the unstructured-PaI setting as FedPaI-U and the structured-PaI setting as FedPaI-S. To further analyze the impact of client-side personalized sparsity patterns, we implement also server-side versions of FedPaI-U, referred to as FedPaI-U (server), for ablation studies of personalized client-side sparsity.

\subsection{Accuracy Evaluation}
\begin{figure*}[htbp]
    \centering
    \includegraphics[width=\textwidth]{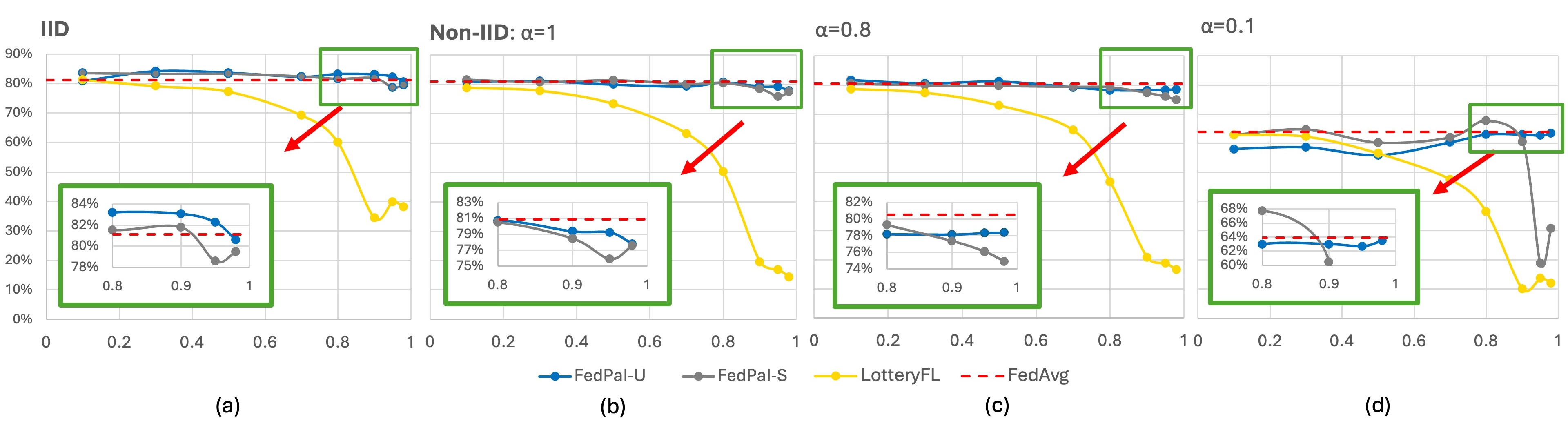} 
    \caption{Accuracy (y-axis) vs. sparsity ratio (x-axis) for IID and non-IID settings of VGG19 model on CIFAR10.}
    \label{fig:accuracy vs. sparsity}
\end{figure*}

To show how FedPaI improves the efficiency of FL while maintaining strong learning ability, we compare the top-1 training accuracy of VGG19 and ResNet18 (in Appendix~\ref{apdx: resnet}) on the CIFAR-10 dataset. We evaluate four settings: the unpruned model (red dashed line as the baseline), FedPaI-U, FedPaI-S, and LotteryFL, across various sparsity levels (from 10\% to 98\%). The results are presented in Fig.~\ref{fig:accuracy vs. sparsity}.

In the \textbf{\textit{IID setting}}, as shown in Fig.~\ref{fig:accuracy vs. sparsity}(a), both FedPaI-U and FedPaI-S maintain accuracy comparable to or even slightly higher than the baseline across a wide range of sparsity levels, since the pruning method acts as a form of regularization that can suppress overfitting. This demonstrates that PaI can effectively preserve model capacity when sparse patterns are carefully selected. Compared to LotteryFL, both structured and unstructured FedPaI show a clear advantage. As the sparsity surpasses 70\%, the accuracy of LotteryFL drops sharply, likely due to its inability to retain the most critical connections. Meanwhile, FedPaI-U consistently achieves slightly higher accuracy than FedPaI-S, particularly at extreme sparsity levels, benefiting from its flexibility in pruning individual weights. However, FedPaI-S still maintains significantly better accuracy than LotteryFL, highlighting the superiority of PaI over conventional iterative pruning in federated learning.

In the \textbf{non-IID setting}, the Dirichlet concentration parameter $\alpha$ determines the degree of data imbalance among clients, with lower values indicating more skewed distributions. To simulate varying levels of heterogeneity in FL scenarios, we experiment with $\alpha$ values of 0.1, 0.8, and 1.0. Our results reveal that the superiority of PaI still persists in non-IID scenarios. Both FedPaI-U and FedPaI-S sustain high accuracy levels close to the unpruned baseline, whereas LotteryFL exhibits a rapid accuracy decline, failing to converge when the pruning rate surpasses 70\% for $\alpha=1$ and 0.8. This confirms that PaI-based approaches are more robust to data heterogeneity, preserving learning capacity even under varying levels of data imbalance. 

For the extremely unbalanced ($\alpha$=0.1) case, as shown in Fig.~\ref{fig:accuracy vs. sparsity}(d), we find that FedPaI-U remains effective in maintaining learning performance, while FedPaI-S starts to degrade at sparsity levels exceeding 95\%. This is because when data is extremely heterogeneous, the optimal connection of each client might be significantly varied, so the FedPaI-U, being a client-side personalized pruning method and allowing each client to adaptively prune its model based on local data characteristics, makes it more resilient to severe non-IID distributions. In contrast, FedPaI-S enforces server-side pruning, which limits personalized exploration of fine-grained structures, ultimately hindering convergence at extreme sparsity levels.

\begin{figure}[tb]
    \centering
    \subfigure{
        \includegraphics[width=0.85\linewidth]{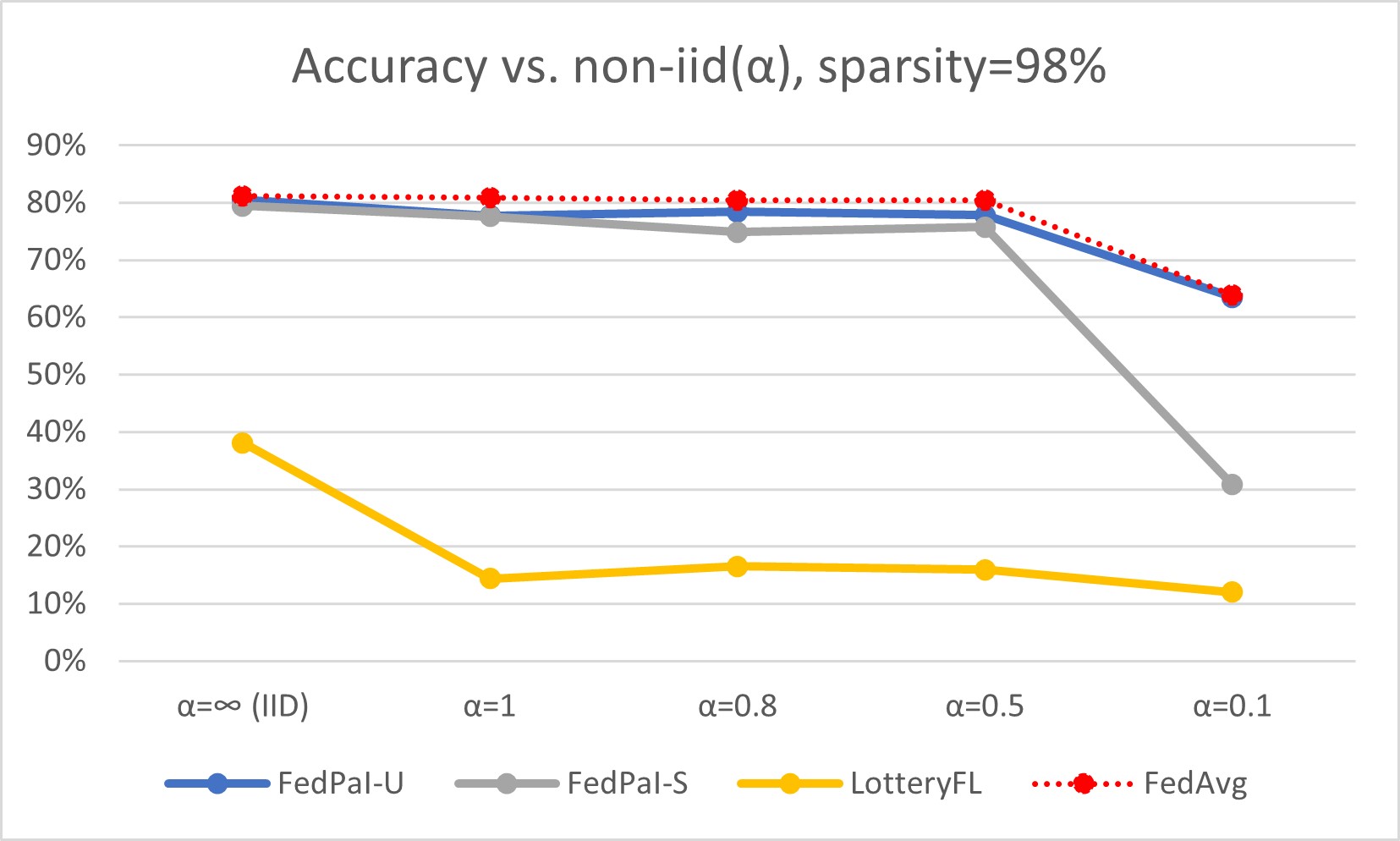}
        \label{fig:acc-vs-alpha}
    }
    \subfigure{
        \includegraphics[width=0.85\linewidth]{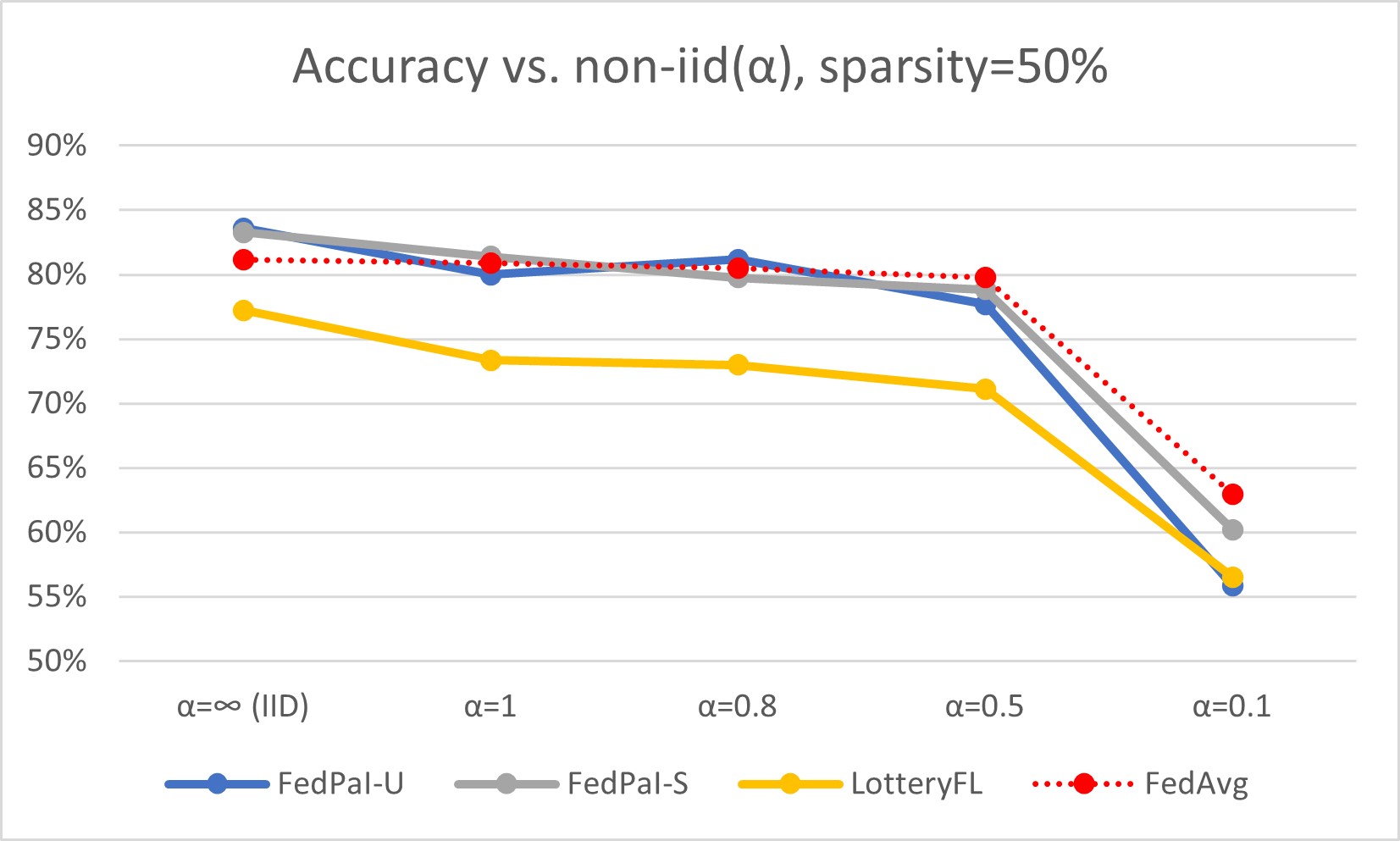}
        \label{fig:acc-vs-alpha-sp50}
    }
    \caption{Comparison of accuracy (y-axis) vs. non-IID level $\alpha$ (x-axis) for the VGG19 model on CIFAR10 under different sparsity levels.}
    \label{fig:acc-vs-alpha-combined}
\end{figure}

\textbf{\textit{Robustness against unbalanced data.}} We also show the trend of accuracy drop as the $\alpha$ decreases in Fig.~\ref{fig:acc-vs-alpha-combined}. 
We can see that under an extreme sparsity of 98\%, FedPaI-U demonstrates remarkable resilience to data heterogeneity, maintaining performance comparable to vanilla FedAvg across different $\alpha$ values. Under highly skewed distributions ($\alpha=0.1$), FedPaI-U significantly outperforms other pruning approaches. In contrast, FedPaI-S exhibits sensitivity to extreme non-IID scenarios but shows better resilience with medium sparsity (50\%), suggesting that structured pruning patterns may struggle to capture diverse feature representations in highly unbalanced settings, but can perform consistently under different sparsity. Notably, the conventional iterative pruning method employed in LotteryFL always shows poor performance across all data distributions under both high and medium sparsity. These results demonstrate that our unstructured PaI approach better preserves model capacity for handling heterogeneous data distributions while maintaining extreme sparsity. A more detailed illustration of the robust training curve of FedPaI is depicted in Appendix~\ref{apdx: robust}.

\begin{table*}[t]
\centering
\begin{tabular}{|l|l|l|l|l|l|c|}
\hline
\textbf{Work} & \textbf{Dataset} & \textbf{Model} & \textbf{IID/non-IID} & \textbf{Compression Method} & \textbf{Compression Rate} & \textbf{Accuracy}\\
\hline 
\hline
\multirow{2}{*}{AdaQuantFL\cite{jhunjhunwala2021adaptive}} & \multirow{2}{*}{CIFAR10} & \multirow{2}{*}{ResNet18} & IID & \multirow{2}{*}{\ding{55} Quantization} & 8.0$\times$ (4-bit)  & 70.02\%\\
 &  &  & $\alpha$ unknown & & 8.0$\times$ (4-bit)  & 22.8\%\\
\hline
FedMPQ\cite{chen2024mixed} & CIFAR10 & ResNet20 & $\alpha=0.1$ & \ding{55} Mixed-Precision Quantization & 6.4$\times$ (avg. 5-bit) & 49.1\% \\
\hline 
\hline
\multirow{2}{*}{EF-RC\cite{wu2023efficient}} & \multirow{2}{*}{CIFAR10} & \multirow{2}{*}{VGG16} & \multirow{2}{*}{$\alpha=1$} & \multirow{2}{*}{\ding{51} Structured Pruning} & 2.7$\times$ (63\% sparse)  & 72.83\%\\
 &  &  &  & & 4.0$\times$ (75\% sparse)  & 40.15\%\\
\hline
\multirow{2}{*}{pFedGate\cite{chen2023efficient}} & \multirow{2}{*}{CIFAR10} & \multirow{2}{*}{LeNet} & IID & \multirow{2}{*}{\ding{51} Structured Pruning} & 2.0$\times$ (50\% sparse)  & 74.12\%\\
 &  &  & $\alpha=0.1$ & & 2.0$\times$ (50\% sparse)  & 72.55\%\\
\hline
\multirow{2}{*}{LotteryFL\cite{9708944}} & \multirow{2}{*}{CIFAR10} & \multirow{2}{*}{VGG19} & IID & \multirow{2}{*}{\ding{55} Unstructured Pruning} &5.0$\times$ (80\% sparse)  & 60.03\%\\
 &  &  & $\alpha=0.1$ & & 2.0$\times$ (50\% sparse)  & 56.52\%\\
\hline
\hline
\multirow{2}{*}{\textbf{FedPaI-U}} & \multirow{2}{*}{CIFAR10} & \multirow{2}{*}{VGG19} & IID & \multirow{2}{*}{\ding{55} Unstructured Pruning} &\textbf{50.0$\times$} (98\% sparse)  & \textbf{80.58}\%\\
 &  &  & $\alpha=0.1$ & & \textbf{50.0$\times$} (98\% sparse)  & \textbf{63.48}\%\\
\hline
\multirow{2}{*}{\textbf{FedPaI-S}} & \multirow{2}{*}{CIFAR10} & \multirow{2}{*}{VGG19} & IID & \multirow{2}{*}{\ding{51} Structured Pruning} & \textbf{50.0$\times$} (98\% sparse)  & \textbf{79.48}\%\\
 &  &  & $\alpha=0.1$ & & \textbf{5.0$\times$} (80\% sparse)  & \textbf{67.71}\%\\
\hline
\end{tabular}
\vspace{0.1in}
\caption{Efficiency vs. accuracy of existing efficient FL works.}
\label{tab:cifar10_fl}
\parbox{\linewidth}{\footnotesize \ding{51} denotes it enables acceleration without dedicated hardware support.}
\end{table*}

\subsection{Ablation Study on Accuracy Gains}
\begin{figure}[tb]
    \centering
    \includegraphics[width=1\linewidth]{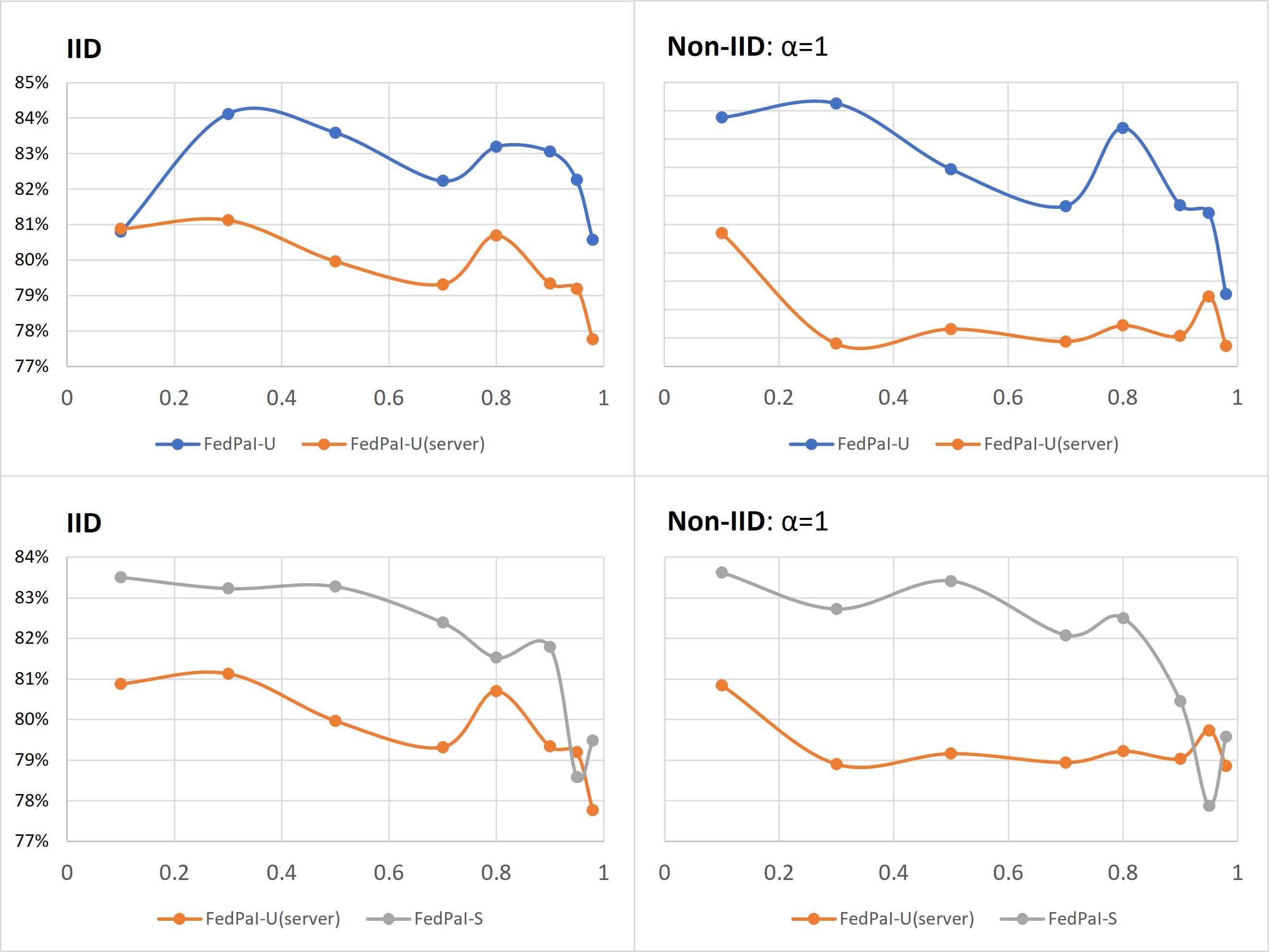}
    \caption{Ablation study between FedPaI-U, FedPaI-S, and FedPaI-U(server).}
    \label{fig:ablation}
\end{figure}
To comprehensively evaluate FedPaI's performance and adaptability, we conduct detailed ablation studies comparing different pruning strategies and their implementations. We examine both unstructured (FedPaI-U) and structured (FedPaI-S), with FedPaI-U incorporating an advanced client-side personalized pruning scheme. To isolate the impact of client-side personalization from the inherent benefits of unstructured pruning, we implement a server-side variant, FedPaI-U(server), for comparison.

As illustrated in Fig.~\ref{fig:ablation}, FedPaI-U consistently outperforms its server-side counterpart by 1-5\% accuracy across both IID and non-IID settings ($\alpha=1$). This significant improvement demonstrates the effectiveness of client-side pruning in capturing client-specific features. Interestingly, when comparing server-side implementations, FedPaI-S exhibits superior performance over FedPaI-U(server). This highlights the effectiveness of EBT's early-stopping mechanism in preserving global features by combining the information of multiple epochs from all clients during the structured pruning. These findings underscore two key insights: (1) the critical importance of client-side personalization in federated settings, as evidenced by FedPaI-U's superior performance, and (2) the necessity of carefully adapting pruning strategies to FL's unique characteristics. Our proposed methods successfully address both aspects, achieving enhanced performance through thoughtful integration of personalization and pruning mechanisms.

\subsection{Efficiency Analysis}
The ultimate goal for exploring pruning in FL is to reduce the resource requirements and further accelerate training while still maintaining good model performance. Although we simulate the FedPaI in a centralized environment with GPU integration, we conduct efficiency analysis by profiling practical deployments by tracking the resource and time consumptions, so that it can reflect the resource reduction and training acceleration if deployed on a real distributed FL infrastructure.

\textit{\textbf{Communication and computation efficiency}}. We evaluate the overall efficiency of the proposed FedPaI from two perspectives: communication requirements and computation overhead. Table~\ref{tab:cifar10_fl} presents a comprehensive comparison between FedPaI and existing efficient FL approaches, highlighting their performance across different compression methods and data distribution settings. Existing quantization-based approaches, such as AdaQuantFL and FedMPQ, achieve notable compression rates (8.0× and 6.4×, respectively). However, these methods exhibit significant performance degradation under non-IID settings, with FedMPQ achieving only 49.1\% accuracy at $\alpha=0.1$. Moreover, these quantization methods require specialized hardware support for low-precision arithmetic to realize computational benefits, limiting their practical deployment. Current pruning-based FL approaches demonstrate moderate compression rates ranging from 2.0× to 5.0×. For instance, pFedGate can only achieve 50\% sparsity, while LotteryFL shows limited robustness to non-IID data, achieving only 56.52\% accuracy at $\alpha=0.1$ with the same sparsity level. 

In contrast, FedPaI demonstrates remarkable efficiency while maintaining superior accuracy. FedPaI-U achieves an unprecedented 50.0× compression rate (98\% sparsity) while maintaining 80.58\% accuracy under IID settings and 63.48\% under extreme non-IID conditions ($\alpha=0.1$). Similarly, FedPaI-S maintains robust performance (79.48\% IID, 67.71\% non-IID) at high sparsity levels. These results underscore FedPaI's superior ability to balance extreme sparsification with model performance, significantly outperforming existing approaches in both efficiency and accuracy metrics.

\begin{figure}[tb]
    \centering
    \begin{subfigure}
        \centering
        \includegraphics[width=0.8\linewidth]{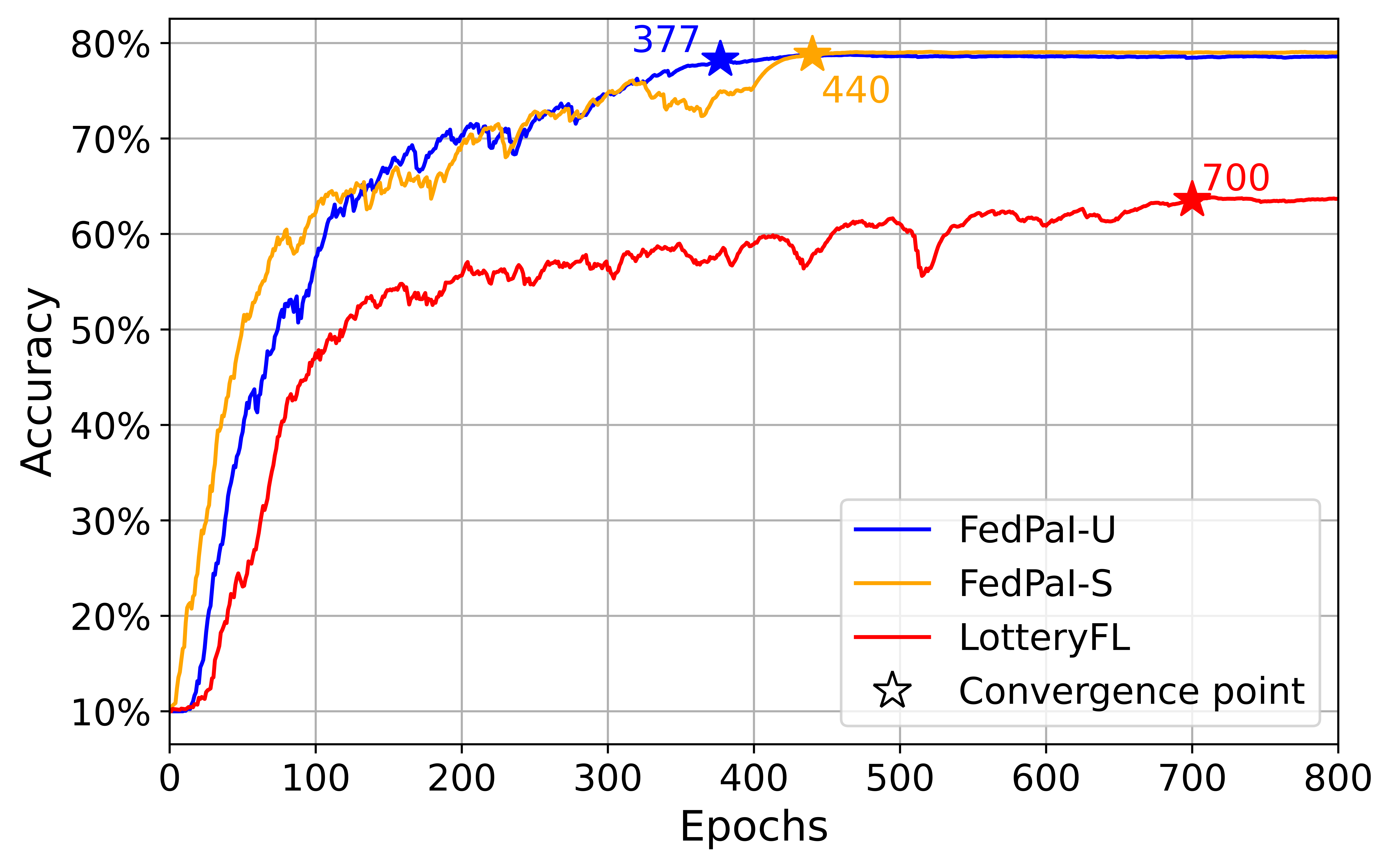}
    \end{subfigure}
    \hfill
    \begin{subfigure}
        \centering
        \includegraphics[width=0.8\linewidth]{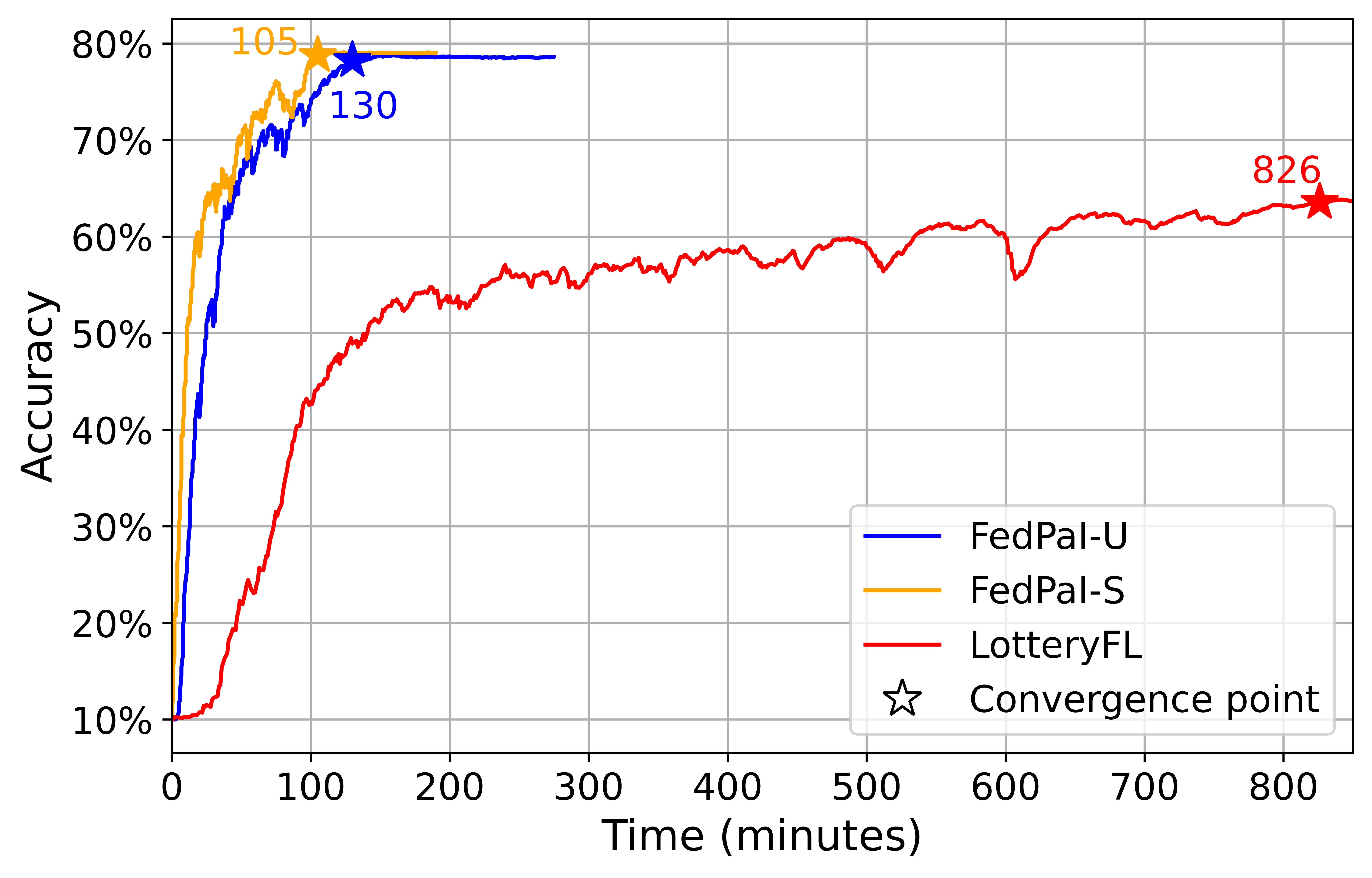}
    \end{subfigure}
    \caption{Comparison of convergence for VGG19 on CIFAR10 under 70\% sparsity and non-IID setting $\alpha=0.8$. (a) Convergence speed, (b) Convergence time. (The curves were smoothed using exponential moving average to reduce noise and enhance visual clarity. )} 
    \label{fig:convergence_combined}
\end{figure}

\textit{\textbf{System level acceleration.}} Although pruning methods can effectively reduce the resource requirement of FL, it is usually non-trivial to convert the sparsity to system-level acceleration. This is because, even though the pruning can lower communication costs for FL training, it may harm the representative ability of the model, not only leading to suboptimal model accuracy but also a slower convergence speed. To illustrate the training speed of each method, We plot the training curve for a general setting of 70\% sparsity and $\alpha=0.8$ w.r.t. the training epoch in Figure~\ref{fig:convergence_combined}(a). We define the convergence point as the epoch at which the test accuracy reaches its maximum or remains stable, and mark them by the star symbol to highlight the difference in convergence speed between different approaches. We can see both FedPaI-U and FedPaI-S significantly outperform the LotteryFL in convergence speed by around 1.6$\times$ to 1.9$\times$. This is because our FedPaI-U can leverage gradient flow information to identify the optimal training connections after pruning, and FedPaI-S detects a structured optimal pruning pattern by collecting global information for the first few epochs. The pruning structure of FedPaI still preserves great learning capacity and, thus, requires fewer epochs to converge. In contrast, LotteryFL, which employs conventional iterative pruning, requires significantly more epochs (up to 1000 in some trials) to reach the convergence point, which in turn cancels out the system-level acceleration brought by its sparsity.

Considering FedPaI-U requires specialized sparse-aware hardware support to unlock its full acceleration potential, it cannot leverage the computation reduction brought by pruning to accelerate training. In order to provide users the flexibility to achieve acceleration with heterogeneous infrastructure, FedPaI-S is a better choice for tangible speedups on standard hardware. We implement the FedPaI-S by actually pruning the channels and only reserve the smaller, pruned model on the device to reduce resource consumption and accelerate training. We measure the actual training time of all methods on the Nvidia A100, and plot the training time in Figure~\ref{fig:convergence_combined}(b). The training time for each communication round of FedPaI-S is 1.5$\times$ faster than FedPaI-U. Besides, since PaI can fix the sparsity pattern at an early stage, while LTH in LotteryFL requires frequent updates of masks and weight, which brings about huge computation overhead, FedPaI-S achieves 5.1$\times$ faster training than LotteryFL for each round. Compared to LotteryFL, our FedPaI can achieve significant acceleration by 6.4$\times$ to 7.9$\times$ in training time on general hardware.

\begin{figure*}[t]
	\centering
	\includegraphics[width=1\linewidth]{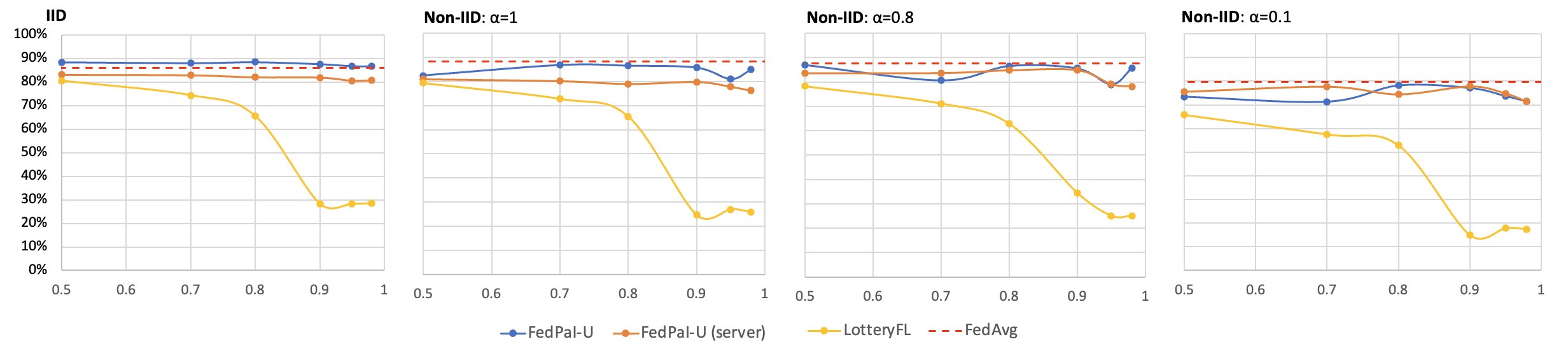}
	\caption{Accuracy (y-axis) vs. sparsity ratio (x-axis) for IID and non-IID settings of ResNet18 model on CIFAR10.}
	\label{fig:accuracy vs. sparsity-resnet}
\end{figure*}

\section{Conclusion}
In this paper, we proposed FedPaI, an FL framework that leverages PaI to achieve extreme sparsity while maintaining high model accuracy. By identifying optimal sparse connections at the start of training, FedPaI reduces communication and computation overhead, outperforming conventional iterative pruning methods. Our results demonstrate that FedPaI achieves up to 98\% sparsity without accuracy degradation, even under challenging non-IID settings, and accelerates training by 6.4$\times$ to 7.9$\times$. With its flexibility in supporting structured and unstructured pruning, FedPaI offers a scalable and efficient solution for diverse FL applications on resource-constrained edge environments.


\bibliographystyle{IEEEtran} 
\bibliography{refs}  

\newpage
\appendix
\subsection{Experiments of ResNet Model}\label{apdx: resnet}
To evaluate the generalizability of the proposed FedPaI across different model architectures, we implement FedPaI using the ResNet18 model and assess its performance. The experimental results, presented in Figure~\ref{fig:accuracy vs. sparsity-resnet}, demonstrate a similar accuracy trend to the VGG experiments. These findings confirm that FedPaI can achieve extremely high sparsity levels with minimal or no degradation in model performance, further validating its effectiveness across diverse architectures.

\subsection{Robust non-IID training with PaI}\label{apdx: robust}
\begin{figure}
    \centering
    \includegraphics[width=0.9\linewidth]{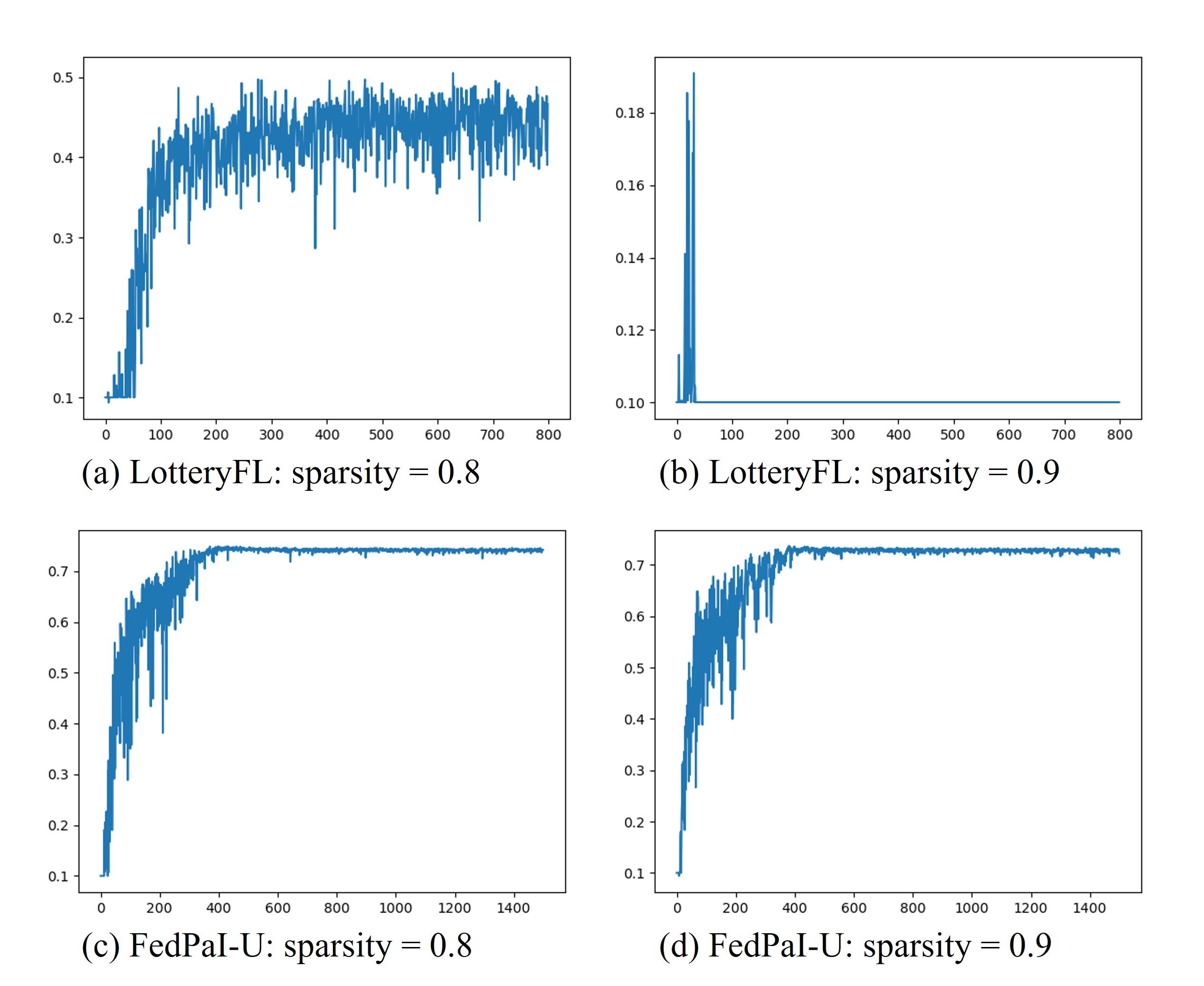}
    \caption{Training curve of FedPaI and LotteryFL (non-IID $\alpha=0.5$).}
    \vspace{-0.2in}
    \label{fig:curves}
\end{figure}

We further illustrate the training curves for LotteryFL and FedPaI-U under the non-IID setting with $\alpha=0.5$, as shown in Figure~\ref{fig:curves}. In non-IID scenarios, existing efficient FL approaches that rely on conventional iterative pruning often experience unstable training, with the instability becoming more pronounced as the pruning rate increases. At extreme sparsity levels, these methods frequently collapse due to their suboptimal structures and limited model capacity. In contrast, FedPaI demonstrates significantly more stable training curves with reduced noise, attributed to its ability to identify optimal connection structures early in the training process, preserving both stability and learning capacity.

\end{document}